\documentclass[a4paper,twoside]{article}

\usepackage{epsfig}
\usepackage{calc}
\usepackage{amssymb}
\usepackage{amstext}
\usepackage{amsmath}
\usepackage{amsthm}
\usepackage{multicol}
\usepackage{pslatex}
\usepackage{apalike}
\usepackage{subfig} 
\usepackage{amssymb}
\usepackage{booktabs}
\usepackage{SCITEPRESS}     

\begin{document}
\title{SAT-MARL: Specification Aware Training in Multi-Agent Reinforcement Learning}


\author{\authorname{Fabian Ritz\sup{1}, Thomy Phan\sup{1}, Robert Müller\sup{1}, Thomas Gabor\sup{1}, Andreas Sedlmeier\sup{1}, Marc Zeller\sup{2}, Jan Wieghardt\sup{2}, Reiner Schmid\sup{2}, Horst Sauer\sup{2}, Cornel Klein\sup{2} and Claudia Linnhoff-Popien\sup{1}}
\affiliation{\sup{1} Mobile and Distributed Systems Group, LMU Munich, Germany}
\affiliation{\sup{2} Corporate Technology (CT), Siemens AG, Germany}
\email{fabian.ritz@ifi.lmu.de}
}


\abstract{A characteristic of reinforcement learning is the ability to develop unforeseen strategies when solving problems. While such strategies sometimes yield superior performance, they may also result in undesired or even dangerous behavior. In industrial scenarios, a system's behavior also needs to be predictable and lie within defined ranges. To enable the agents to learn (how) to align with a given specification, this paper proposes to explicitly transfer functional and non-functional requirements into shaped rewards. Experiments are carried out on the \textit{smart factory}, a multi-agent environment modeling an industrial lot-size-one production facility, with up to eight agents and different multi-agent reinforcement learning algorithms. Results indicate that compliance with functional and non-functional constraints can be achieved by the proposed approach.}

\onecolumn \maketitle \normalsize \setcounter{footnote}{0} \vfill

\section{Introduction}

\emph{Reinforcement learning (RL)} enables an autonomous agent to optimize its behavior even when its human programmers do not know what optimal behavior looks like in a specific situation~\cite{sb18}. Recent breakthroughs have shown that RL even allows agents to surpass human performance~\cite{silver2017mastering}. Yet, RL systems are typically given a well-defined target, e.g. ``win the game of chess''. But when applying RL to real-world problems such as industrial applications, the ideal target itself is often less clear. In this paper, we consider an adaptive production line as it is supposed to be part of the factory of the near future~\cite{wang2016towards}.

Typically, a smart factory is given a clear \emph{functional requirement} in the form of an order of items that can be translated into a series of tasks for the present machines~\cite{step18}. Its goal is to produce all ordered items within a certain time frame. However, there often exist a lot of \emph{non-functional requirements} as well: the system should not exhaust all of its time if faster production is possible; it should avoid operations that could damage or wear down the machines; it should be robust to unexpected events and human intervention~\cite{cheng2009software,belzner2016software,bures2017software}. The full set of  requirements is the \emph{specification} of a system.

The fulfillment of a given specification could be regarded as a clear target for an RL agent. However, it involves an intricate balance of achieving the convoluted requirements at the same time, resulting in a sparse reward signal that prohibits any learning progress. Still, weighing various requirements while neither introducing erratic nor unsafe behavior was also discovered to be difficult challenge in the literature~(see Sec.~\ref{sec:related-work}). E.g., if multiple agents need to leave through the same exit, approaches such as restricting single actions might prevent collisions but would not incentivize the agents to learn coordinated behavior, thus resulting in a deadlock. In Sec.~\ref{sec:domain}, we introduce a new domain based on a smart factory populated by multiple agents acting independently. For this setting, we show how a full specification of functional and non-functional requirements can be transferred into reward functions for RL (see Sec.~\ref{sec:transfer}). Evaluating the different reward schemes in Sec.~\ref{sec:evaluation}, we observe that some non-functional requirements are (partially) subsumed by overarching functional requirements, i.e. they are learned easily, while others significantly affect convergence and performance. Our main contributions are:

\begin{itemize}
	\item A novel multi-agent domain based on the industrial requirements of a smart factory
	\item The application of specification-driven reward engineering to a multi-agent setting
	\item A thorough evaluation of the impact of typical secondary reward terms on different multi-agent reinforcement learning algorithms
\end{itemize}

\section{Foundations}
\label{sec:foundations}

\subsection{Problem Formulation}

We formulate our problem as \emph{Markov game} $M_{\textit{SG}} = \langle \mathcal{D},\mathcal{S},\mathcal{A},\mathcal{P},\mathcal{R},\mathcal{Z},\Omega \rangle$, where $\mathcal{D} = \{1,...,N\}$ is a set of agents, $\mathcal{S}$ is a set of states $s_{t}$, $\mathcal{A} = \mathcal{A}_{1} \times ... \times \mathcal{A}_{N}$ is the set of joint actions $a_{t} = \langle a_{t,1},...,a_{t,N} \rangle$, $\mathcal{P}(s_{t+1}|s_{t}, a_{t})$ is the transition probability, $r_{t,i} = \mathcal{R}_{i}(s_{t},a_{t})$ is the reward of agent $i \in \mathcal{D}$, $\mathcal{Z}$ is a set of local observations $z_{t,i}$ for each agent $i$, and $\Omega(s_{t},a_{t}) = z_{t+1} = \langle z_{t+1,1},...,z_{t+1,N} \rangle \in \mathcal{Z}^{N}$ is the joint observation function. For cooperative MAS, we assume a common reward function $r_{t} = \mathcal{R}(s_{t},a_{t})$ for all agents.

The behavior of a MAS is defined by the (stochastic) \emph{joint policy} $\pi(a_{t}|z_{t}) = \langle\pi_{1}(a_{t,1}|z_{t,1}),...,\pi_{N}(a_{t,N}|z_{t,N})\rangle$, where $\pi_{i}(a_{t,i}|z_{t,i})$ is the \emph{local policy} of agent $i$.

The goal of each agent $i$ is to find a \emph{local policy} $\pi_{i}(a_{t,i}|z_{t,i})$ as probability distribution over $\mathcal{A}_{i}$ which maximizes the expected discounted local return or local action value function $Q_{i}(s_{t},a_{t}) = \mathbb{E}_{\pi}[\sum_{k=0}^{\infty} \gamma^{k} \mathcal{R}_{i}(s_{t+k}, a_{t+k}) \; | \; s_{t},a_{t}]$, where $\pi = \langle \pi_{1},...,\pi_{N} \rangle$ is the \emph{joint policy} of all agents and $\gamma \in [0,1]$ is the discount factor. The optimal local policy $\pi_{i}^{*}$ of agent $i$ depends on the local policies $\pi_{j}$ of all other agents $j \neq i$.

\subsection{Multi-Agent Reinforcement Learning}
In \emph{Multi-Agent Reinforcement Learning (MARL)}, each agent $i$ searches for an (near-)optimal local policy $\pi_{i}^{*}(a_{t,i}|z_{t,i})$, given the policy $\pi_{j}$ of all other agents $j \neq i$. In this paper, we focus on value-based RL. Local action value functions are commonly represented by deep neural networks like \emph{DQN} to solve high-dimensional problems \cite{mnih2015human,silver2017mastering}.

DQN has already been applied to MARL, where each agent is controlled by an individual DQN and trained independently \cite{leibo2017multi,tampuu2017multiagent}. While independent learning offers scalability w.r.t. the number of agents, it lacks convergence guarantees, since the adaptive behavior of each agent renders the system non-stationary, which is a requirement for single-agent RL to converge \cite{laurent2011world}.

Recent approaches to MARL adopt the paradigm of \emph{centralized training and decentralized execution (CTDE)} to alleviate the non-stationarity problem \cite{rashid2018qmix,sunehag2017value,son2019qtran}. During centralized training, global information about the state and the actions of all other agents are integrated into the learning process in order to stably learn local policies. The global information is assumed to be available, since training usually takes place in a laboratory or in a simulated environment.

Value decomposition or factorization is a common approach to CTDE for value-based MARL in cooperative MAS. \emph{Value Decomposition Networks (VDN)} are the most simple approach, where a linear decomposition of the global $Q^{*}(s_{t},a_{t})$-function is learned to derive individual value functions for each agent \cite{sunehag2017value}. Alternatively, the individual value functions of each agent can be mixed with a non-linear function approximator to learn the $Q^{*}(s_{t},a_{t})$-function \cite{rashid2018qmix,son2019qtran}. \emph{QMIX} is an example for learning non-linear decompositions which uses a monotonicity constraint, where the maximization of the global $Q^{*}(s_{t},a_{t})$-function is equivalent to the maximization of each individual value function \cite{rashid2018qmix}.

\subsection{Reward Shaping in RL}
Reward shaping is an evident approach to influence an RL agent's behavior.
Specifically, potential based reward shaping (PBRS) \cite{pbrs99} was proven not to alter the optimal policy in single-agent systems and not introducing additional side-effects that would allow reward hacking.
In PBRS, the actual reward applied for a time step is the difference between the prior and the posterior state's potential: 
\begin{align}
	F(s,s') & = \gamma \cdot \Phi(s') - \Phi(s)
\end{align}

Initially requiring a static potential function, the properties of PRBS were later shown to hold for dynamic potential functions as well~\cite{dk12}. Subsequently, PBRS has also been theoretically analyzed in and practically applied to MAS and episodic RL~\cite{dyk14,grz17}. The fundamental insight is that PBRS does not alter the Nash equilibria of MAS, but may affect performance in any direction, depending on scenario and applied heuristics. This paper's reward shaping differs from PBRS in one detail: It uses $\gamma=1.0$ during reward shaping while the learning algorithms use $\gamma=0.95$ in most experiments. For the theoretic guarantees of PBRS to hold, learning algorithm and reward shaping must use the same value of $\gamma$. Note that, however, the results of the fourth evaluation scenario (see Sec.~\ref{sec:evaluation}) indicate that the practical impact is negligible (at least in our case) and there is currently no proven optimality guarantee for DQN-based algorithms using deep neural networks for function approximation anyway.

\section{Related Work}
\label{sec:related-work}

Regarding safety in RL, prior work compiled a list of challenges for learning to respect safety requirements in RL~\cite{ais16} and provided a set of gridworld domains allowing to test a single RL agent for safety~\cite{sgw17}. Yet, the fundamental issues remain unsolved. Further, a comprehensive overview of safe RL approaches subdivides the field into modeling either safety or risk~\cite{gf15}. While some approaches use these concepts to constrain the MDP to prevent certain actions, this paper does not model risk or safety explicitly. Instead, it aims for agents learning (how) to align to a given specification as constraining the MDP may become infeasable in complex multi-agent systems.

Regarding learned safety in RL, one recent approach extends the MDP by a function mapping state and action to a binary feedback signal indicating the validity of the taken action~\cite{HAL9000}. A second neural network is trained to predict this validity in addition to training a DQN. The DQN’s training objective is augmented by an auxiliary loss pushing Q-values of forbidden actions below those of valid actions. Similarly, another  approach accompanies the Q-Network with an Action Elimination Network (AEN) that is trained to predict the feedback signal~\cite{zhm18}. A linear contextual bandit facilitates the features of the penultimate layer of the AEN and eliminates irrelevant actions with high probability, therefore directly altering the action set. Both approaches reduced certain actions and improved performance, but were evaluated in single agent domains only and not compared with reward shaping.

Regarding the training of MARL system, population-based approaches such as \textit{FTW} enable individuals to learn from internal, dense reward signals complementing the sparse, delayed global reward~\cite{jcw19}. Distinct environment signals are used in handcrafted rewards and a process during that agents learn to optimize the internal rewards in accordance with maximizing the global reward. Similrly, separate discounts  can be used to individually adjusts dense, internal rewards to optimize the top level goal~\cite{llm19}. While these approaches demonstrated the capabilities of shaped, dense rewards in MARL and how automatically evolved rewards can outperform hand-crafted rewards, they only aim on boosting the system's performance and do not consider additional specification constraints. While in (video) games, MARL lead to innovative strategies of which humans were unaware before~\cite{silver2017mastering}, the goal of specification compliance is to avoid unintended side-effects which is especially important for industrial and safety-critical domains~\cite{belzner2016software,bures2017software}.

Reward shaping has also been applied to improve cooperation in MAS by addressing the credit assignment problem, where all agents observe the same reward signal. For example, Kalman filtering can be used to extract individual reward signals from the common reward~\cite{chang2004all}. Moreover, the difference between the original reward and an alternative reward where the agent would have chosen a default action can be used to derive an individual training signal for each agent~\cite{wolpert2002optimal}. Also, counterfactual baselines can be used to improve credit-assignment in policy gradient algorithms~\cite{foerster2018counterfactual}. While these approaches address the problem of improving cooperation, they do not explicitly address non-functional requirements to avoid side-effects.

\section{Smart Factory Domain}
\label{sec:domain}

\begin{figure*}[ht]
     \subfloat[smart factory overview\label{fig:sf_cpps3}]{%
         \includegraphics[width=0.48\textwidth]{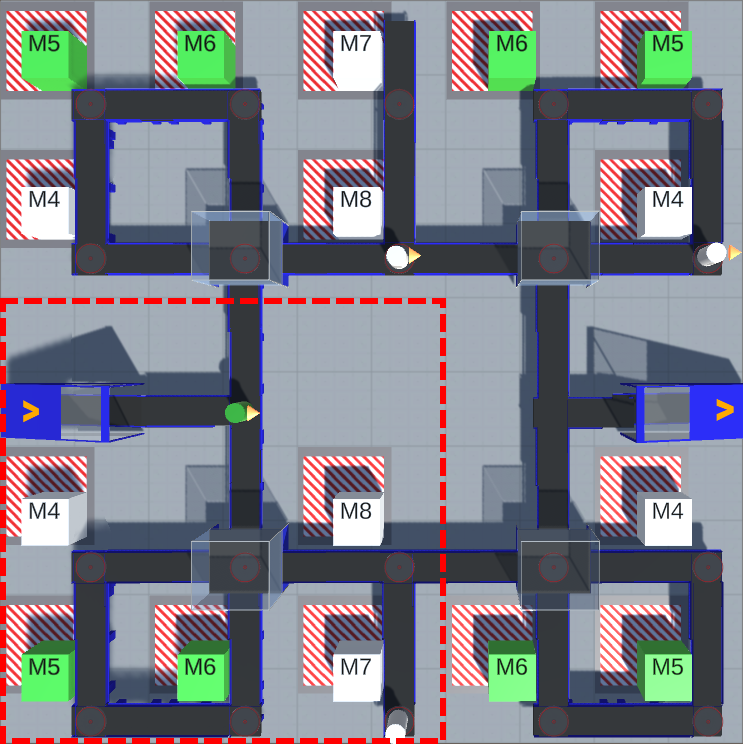}
     }
     \hfill
     \subfloat[zoom-in on bottom left\label{fig:sf_cpps3_zoomed}]{%
         \includegraphics[width=0.48\textwidth]{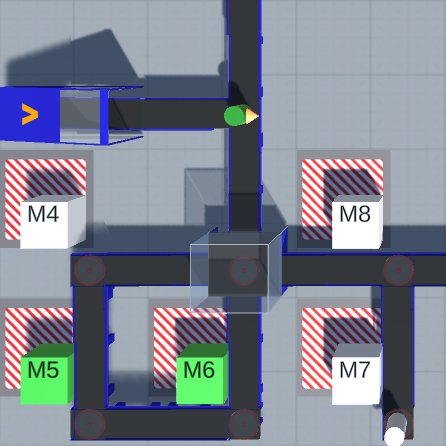}
     }
     \caption{3D visualization of the discrete domain used in this paper. Contrary to a fully connected grid world, the agents can only move to adjacent cells via a connecting path and can only doge each other via on one of the 4 capacitive cells (rendered with transparent boxes). Agents spawn on the entry position on the left and finish on the exit position on the right (both rendered with blue boxes).}
     \label{fig:sf_layouts}
\end{figure*} 

This paper's \textit{smart factory} is inspired by a domain proposed in the literature~\cite{step18} with respect to modeling the components and production processes of a highly adaptive facility in a virtual simulation. It consists of a $5 \times 5$ grid of \emph{machines} with different machine types as shown in Fig.~\ref{fig:sf_layouts}. Each \emph{item} is carried by one agent $i$ and needs to be processed at various machines according to its randomly assigned processing tasks $tasks_{i} = [\{a_{i,1},b_{i,1}\}, \{a_{i,2},b_{i,2}\},...]$, where each task $a_{i,j}$,$b_{i,j}$ is contained in a \emph{bucket}. Fig.~\ref{fig:sf_cpps3_zoomed} shows an example for an agent $i$, rendered as a green cylinder, with $tasks_{i} = [\{5,6\}]$, rendered as green boxes. While tasks in the same bucket can be processed in any order, buckets themselves have to be processed in a specific order. Consequently, $i$ can choose between different machines to process the tasks $a_{i,1} = 5$ and $b_{i,2} = 6$. The agent's initial position is fixed (Fig.~\ref{fig:sf_cpps3}). The agent can move along the machine grid (\emph{left, right, up, down}), \emph{enqueue} at the current position or stay put (\emph{no-op}). The domain is discrete in all aspects including agent motion. Each machine can process exactly one item per time step. Enqueued agents are unable to perform any actions. If a task is processed, it is removed from its bucket. If a bucket is empty, it is removed from the item's tasks list. An item is \emph{complete} if its tasks list is empty.

Contrary to fully-connected grid worlds, every grid cell in the present smart factory has an agent capacity limit and defined connections (paths) to the surrounding grid cells. An agent may only move to another grid cell if a connecting paths exists and the target grid cell's maximum capacity is not exceeded. Entry and exit can hold all agents simultaneously, the four grid cells fully connected to their neighbors can hold half the agents. Fig.~\ref{fig:sf_cpps3_zoomed} shows such a grid cell, located south to example agent $i$, rendered with a transparent box. All other grid cells can only hold one agent. In presence of multiple agents, coordination is required to avoid conflicts when choosing appropriate paths and machines.

\section{Transferring Specification Constraints}
\label{sec:transfer}

\begin{table*}
    \vspace{2mm}
	\caption{reward overview}
	\label{tab:reward_overview}
	\subfloat[reward components: variables and signs\label{tab:component_substitutions}]{
  		\begin{tabular}{lcc}
    	\toprule
    	reward component & variable & sign\\
    	\midrule
    	item completion rew.      & $\alpha$ & +\\
   		single task reward        & $\beta$  & +\\
    	step cost                 & $\delta$ & -\\
    	machine operation cost    & $\zeta$  & -\\
    	path violation penalty    & $\eta$   & -\\
    	agent collision penalty   & $\theta$ & -\\
    	emergency violat. pen.    & $\iota$  & -\\
  		\bottomrule
		\end{tabular}
	}
	\hfill
	\subfloat[reward schemes: variable values\label{tab:component_values}]{
		\begin{tabular}{lccccccc}
		\toprule
    	scheme & $\alpha$ & $\beta$ & $\delta$ & $\zeta$ & $\eta$ & $\theta$ & $\iota$\\
    	\midrule
    	$r0$   & $5.0$ &       & $0.1$ &       &       &       &       \\
        $r1$   &       & $1.0$ & $0.1$ &       &       &       &       \\
        $r2$   &       & $1.0$ & $0.1$ & $0.2$ &       &       &       \\
        $r3$   &       & $1.0$ & $0.1$ &       & $0.1$ &       &       \\
        $r4$   &       & $1.0$ & $0.1$ &       &       & $0.4$ &       \\
        $r5$   &       & $1.0$ & $0.1$ & $0.2$ & $0.1$ & $0.4$ & $1.0$ \\
        $rx$   &       & $1.0$ & $0.1$ &$0-0.2$& $0.1$ &$0-0.4$&$0-1.0$\\
		\bottomrule
		\end{tabular}
	}
	\vspace{-4mm}
\end{table*}

Inspired by PBRS, this paper proposes to omit primary rewards, transfer both functional and non-functional requirements into a potential function and use potential differences as rewards. Setting the functional domain goal to complete items as fast as possible, a first approach is to increase the potential by $\alpha$ once an agent completes its item and decrease it by $\delta$ per step. This is implemented by reward scheme~\textit{r0}. 
As positive feedback in~\textit{r0} is sparse and delayed, a decomposition into more dense, positive terms $\beta$, added whenever a single task is finished, may improve learnability. This is implemented by reward scheme~\textit{r1}.

Given an industrial background, the system may need to comply with a certain non-functional specification, resulting in behavioral constraints. In this domain, the evaluated \emph{soft constraints} are to only use the machine type needed by the task, to stay on the defined paths and not to collide with other agents. In the simulation, items processed by wrong machines remain unaltered and any agent trying to move to a grid cell without sufficient capacity or path connection stays put. Therefore, \emph{soft constraints} do not oppose the goal of completing items (fast) as agents would not benefit from violations anyway. Moreover, agents shall freeze if the emergency signal is active as a \emph{hard constraint}. In the simulation, the agents can ignore the emergency signal in order to finish their tasks faster. Therefore, the \emph{hard constraint} introduces a target conflict. As constraint violations shall be minimized, they are transferred into negative terms $\zeta$ (machine operation cost), $\eta$ (path violation penalty), $\theta$ (agent collision penalty) and $\iota$ (emergency violation penalty) of different quantity, considered in the reward schemes \emph{r2}, \emph{r3}, \emph{r4} and \emph{r5}.

Inspired by curriculum learning~\cite{cl09}, reward scheme~\textit{rx} only contains $\beta$, $\delta$ and $\eta$ in the first part of the training process in order to learn the basic task. $\zeta$, $\theta$ and $\iota$ are added later during training, so that some constraints are introduced to the agents gradually.
A summary of all reward schemes, their components and values is given in Tab.~\ref{tab:reward_overview}. To actually employ the reward schemes, the \emph{smart factory} provides a corresponding interface for each component, e.g. the number of \textit{completed items}, at any time step.
Depending on the learning algorithm, the potential function is evaluated either per agent or globally.
Bringing all together:
\begin{align}
	\Phi(s) & = \alpha \cdot \textit{itemCompleted}(s)   + \beta  \cdot \textit{tasksFinished}(s)  \nonumber \\
			& + \delta \cdot \textit{stepCount}(s)\qquad + \zeta  \cdot \textit{machinesUsed}(s)   \nonumber \\
			& + \eta   \cdot \textit{pathViolations}(s)  + \theta \cdot \textit{agentCollisions}(s)\nonumber \\
			& + \iota  \cdot \textit{emergencyViolations}(s)\nonumber
\end{align}

\section{Evaluation}
\label{sec:evaluation}

\subsection{Experimental Setup}

The reward schemes listed in Table~\ref{tab:component_values} were evaluated on different layouts of the smart factory domain. The reported layout (see Fig.~\ref{fig:sf_cpps3}) turned out to be most challenging. Agents always spawn on an entry (on the left), should then process two buckets of each two random tasks and finally move to the exit (the mirror position to the entry). Machines are not grouped by type as one might expect in a real world setting but distributed equally to maintain solvability in presence of up to $8$ agents. Episode-wise training is carried out for 5000 episodes, each limited to $50$ steps. While the components of reward schemes \emph{r0} to \emph{r5} remain fixed during training, \textit{rx} alters $\zeta$, $\theta$ and $\iota$ during training. After adding or altering values, the exploration rate is set back to $0.25$ and the optimizer momentum is reset.

As a \emph{white-box} test, independent DQN is trained in each scenario: due to the individual rewards, agents are able to directly associate the shaped feedback signals with their actions. The DQN consists of two dense layers of $64$ and $32$ neurons, using ELU activation. The output dense layer consists of $|\mathcal{A}_{i}| = 6$ neurons with linear activation. ADAM is used for optimization. Except evaluation scenario 4, Q-values are discounted with $\gamma=0.95$. $\epsilon-greedy$ exploration with linear decay lasts for approx. $1000$ steps. The experience buffer holds up to $20000$ elements. The target network is updated after each $4000$ training steps. Per training step, a batch of $64$ elements is sampled via prioritized experience replay.

As a \emph{black-box} test, VDN and QMIX agents were trained: due to the global reward, agents cannot directly associate the shaped feedback signal with their individual actions (at least in the beginning of the training).
Both VDN and QMIX use the same hyperparameters as DQN and the same architecture for their local Q-networks.
In addition, QMIX uses a mixing network with one hidden, dense layer of $64$ neurons using ELU activation and an output dense layer with a single neuron using linear activation.

\textit{Performance} is captured through \emph{steps until solved}, representing the episode step in which all agents have finished their tasks. By this paper's definition, a lower value indicates better performance. \textit{Compliance} is measured in \emph{soft} and \emph{hard constraint violations}. These are summed over all agents and all steps of an episode. Again, by definition, lower values indicate higher compliance. Depending on potential function and learning algorithm, these metrics may not always be fully visible to the agents. For all values, mean and $95\%$ confidence interval of $10$ independently trained networks are reported. Experiments are structured in four scenarios:
\begin{enumerate}
	\item To analyze the impact of isolated reward components on compliance and performance during training, reward schemes \textit{r2}, \textit{r3} and \textit{r4} are applied on $4$ DQN agents. For comparison, reward scheme \textit{r1} is always evaluated. The sparse reward scheme \textit{r0} and the combined reward scheme \textit{r5} are evaluated in an additional overview.
	\item To quantify the impact of combined reward components on scalability, reward scheme \textit{r5} is applied on $8$ agents with DQN, VDN and QMIX and compared with reward scheme \textit{r1}.
	\item To examine whether scalability can be improved, reward scheme \textit{rx} is gradually applied on $8$ agents with DQN, VDN and QMIX and compared to the static reward schemes \textit{r1} and \textit{r5}.
	\item To outline whether the proposed approach could be used in safety-critical domains, reward scheme \textit{rx} is compared to \textit{r5} during the training of $6$ DQN agents in a scenario with emergency signals that introduce target conflicts. To not break with the theoretical guarantees of PBRS in this particular scenario, DQN discounts with $\gamma=1.0$ and agents are moved to an absorbing state with zero potential at the end of each episode as proposed in the literature~\cite{dyk14}, thus the number of steps peaks at $51$.
\end{enumerate}

\subsection{Results}

\begin{figure*}
	\captionsetup[subfigure]{justification=justified}
    \subfloat[compliance when puni-\newline shing wrong enqueueing\label{fig:ca_1}]{%
        \includegraphics[trim={0 0 30 30}, clip, width=0.24\textwidth]{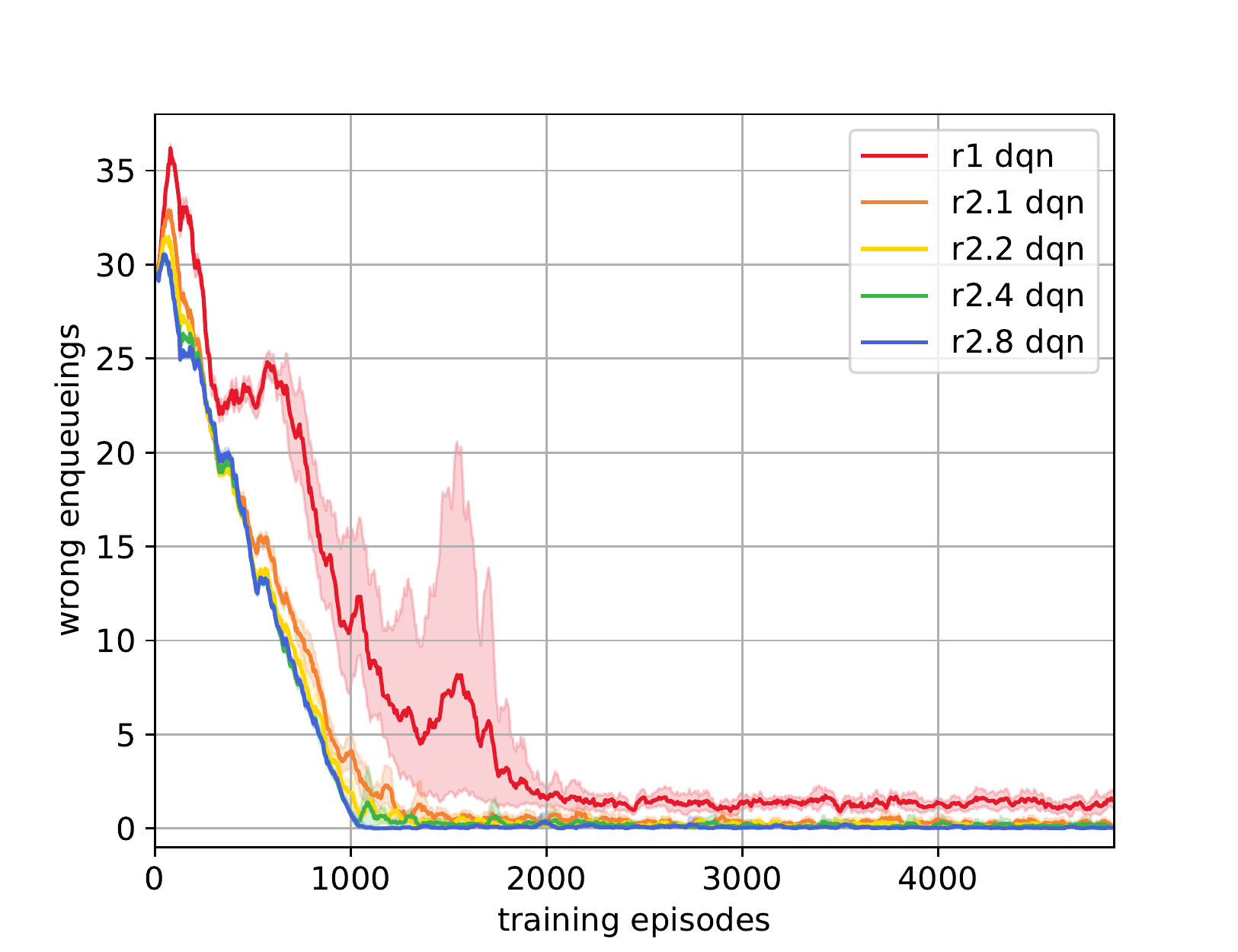}
    }
    \hfill
    \subfloat[compliance when puni-\newline shing path violations\label{fig:ca_2}]{%
        \includegraphics[trim={0 0 30 30}, clip, width=0.24\textwidth]{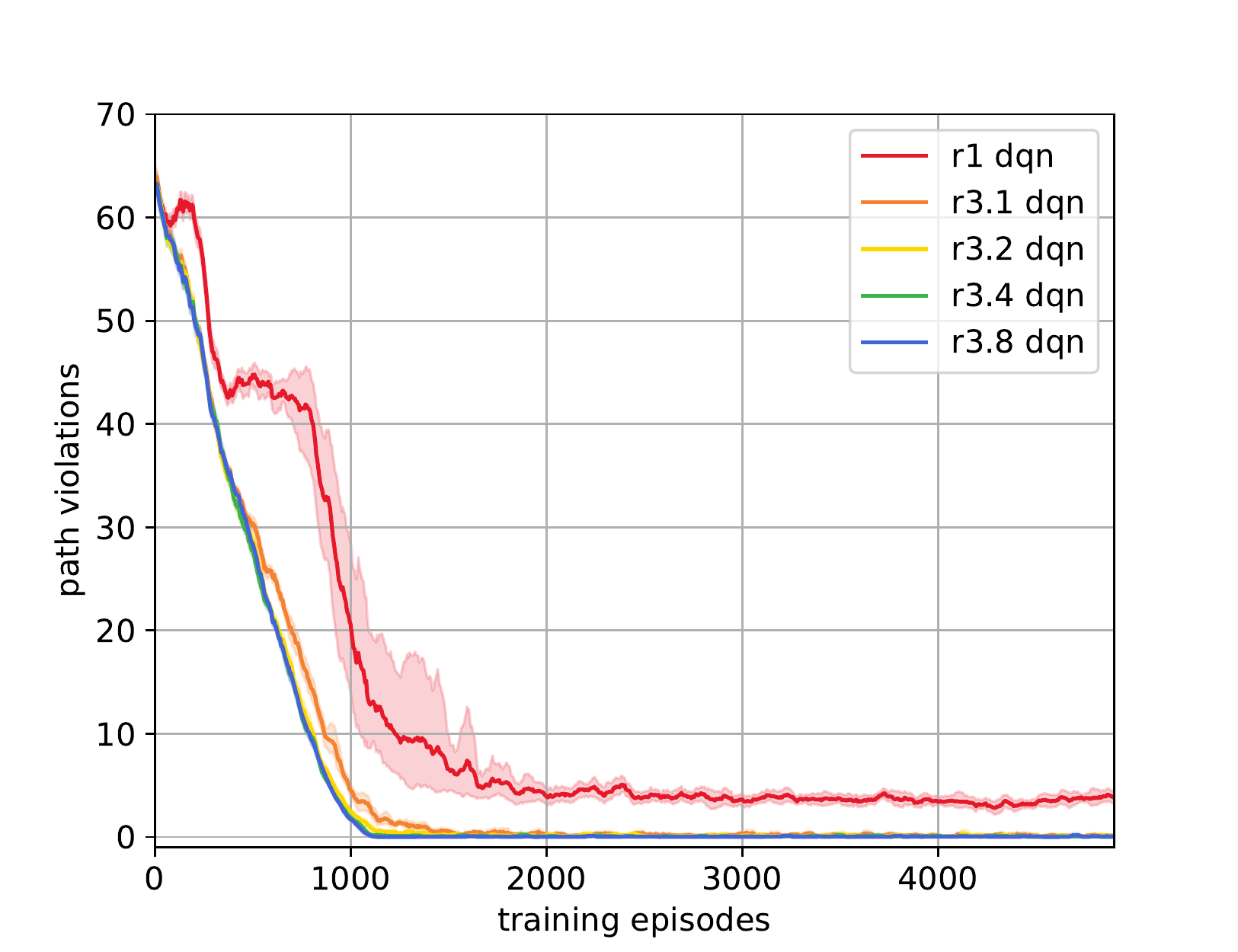}
    }
    \hfill
    \subfloat[compliance when puni-\newline shing agent collisions\label{fig:ca_3}]{%
        \includegraphics[trim={0 0 30 30}, clip, width=0.24\textwidth]{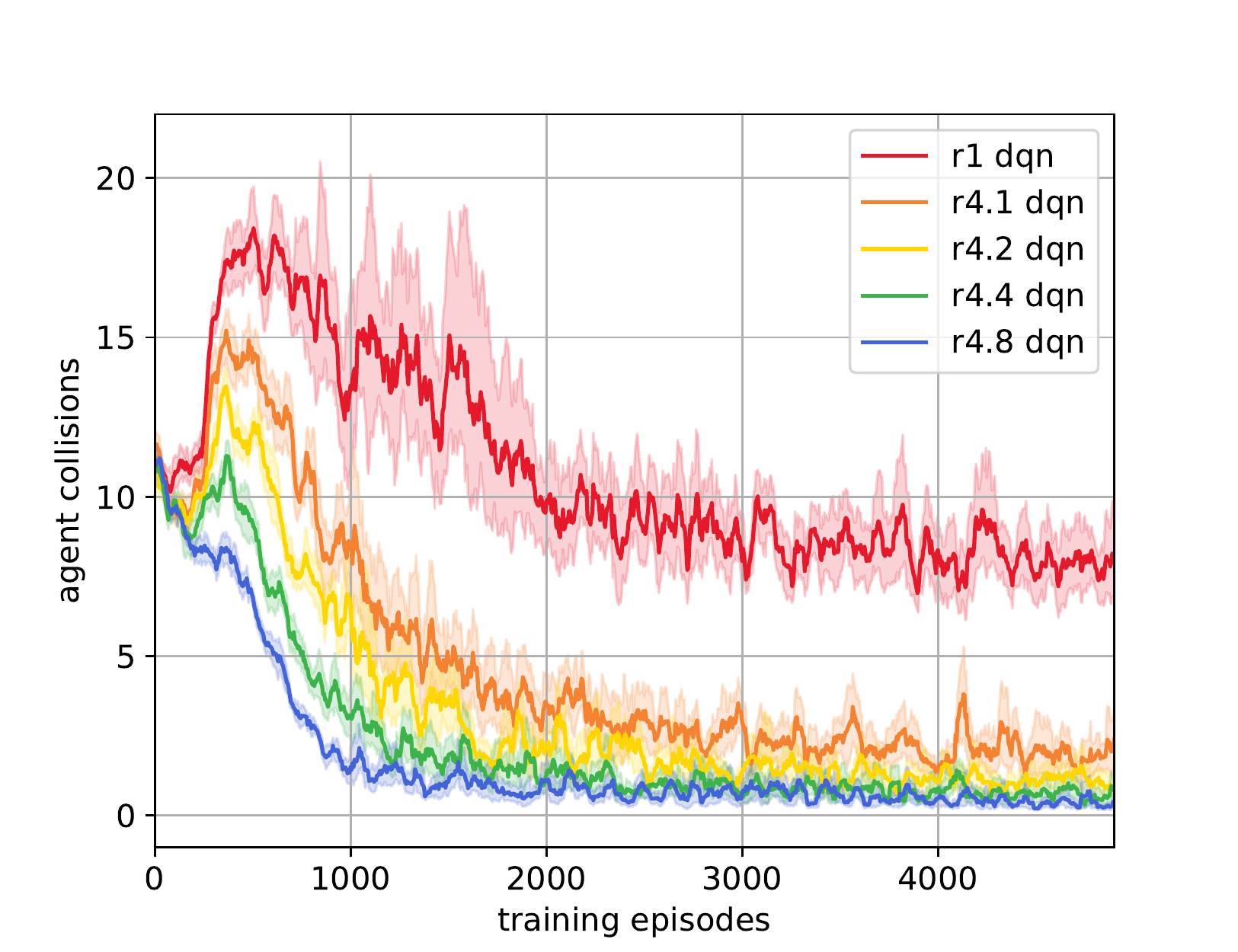}
    }
    \hfill
    \subfloat[compliance overview of all static reward schemes\label{fig:ca_0}]{%
        \includegraphics[trim={0 0 30 30}, clip, width=0.24\textwidth]{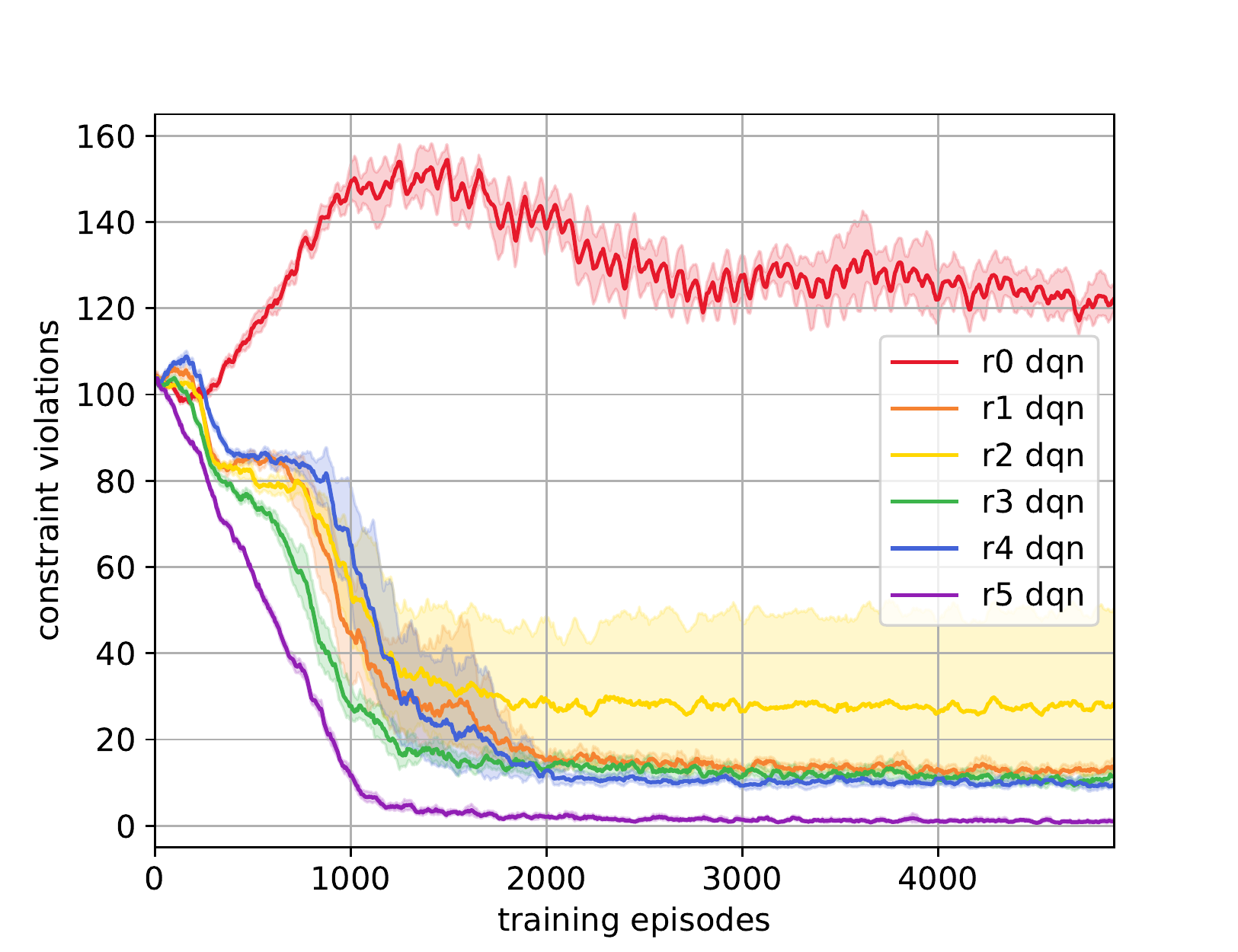}
    }
    \\
    \subfloat[performance when puni-\newline shing wrong enqueueing\label{fig:ca_5}]{%
        \includegraphics[trim={0 0 30 30}, clip, width=0.24\textwidth]{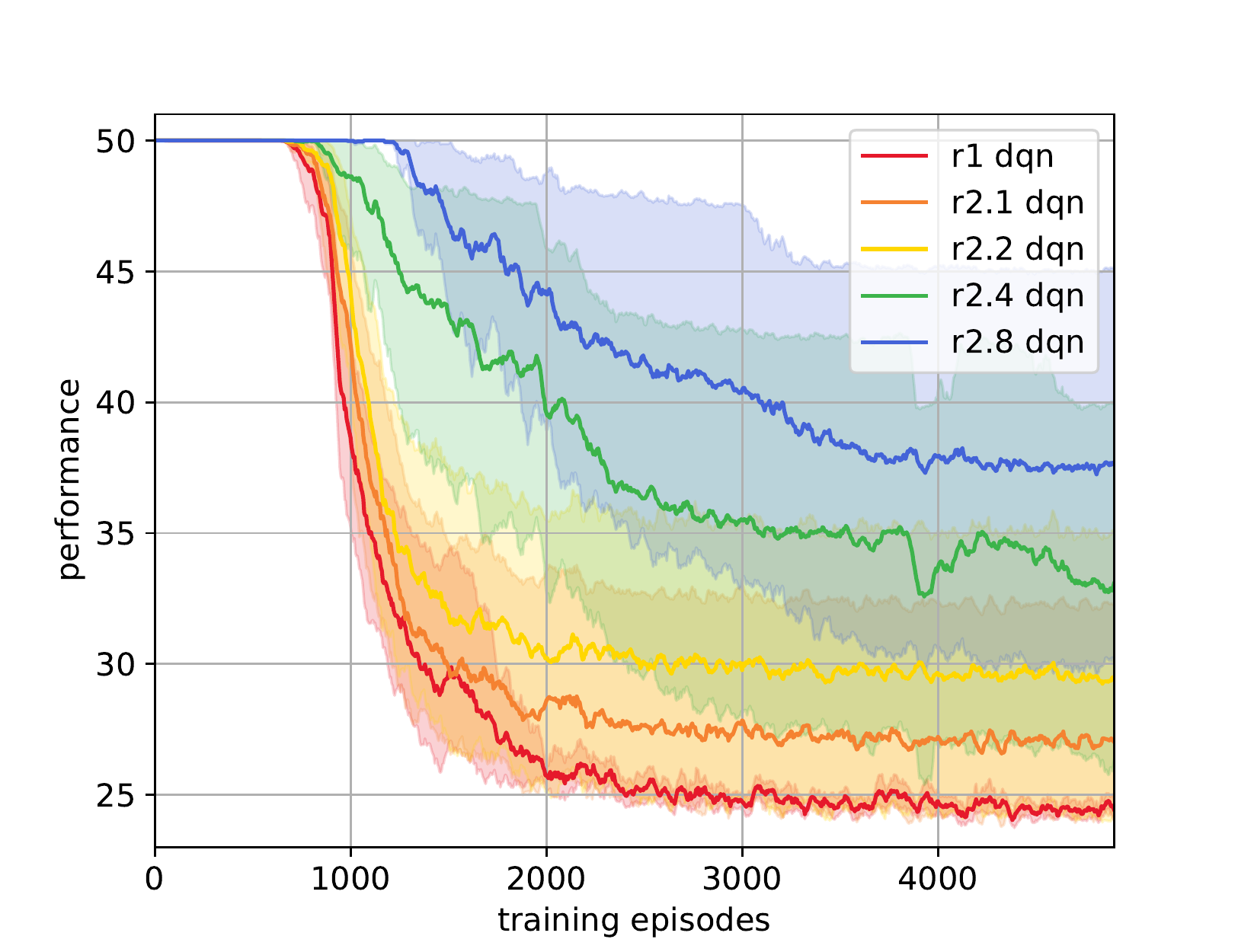}
    }
    \hfill
    \subfloat[performance when puni-\newline shing path violations\label{fig:ca_6}]{%
        \includegraphics[trim={0 0 30 30}, clip, width=0.24\textwidth]{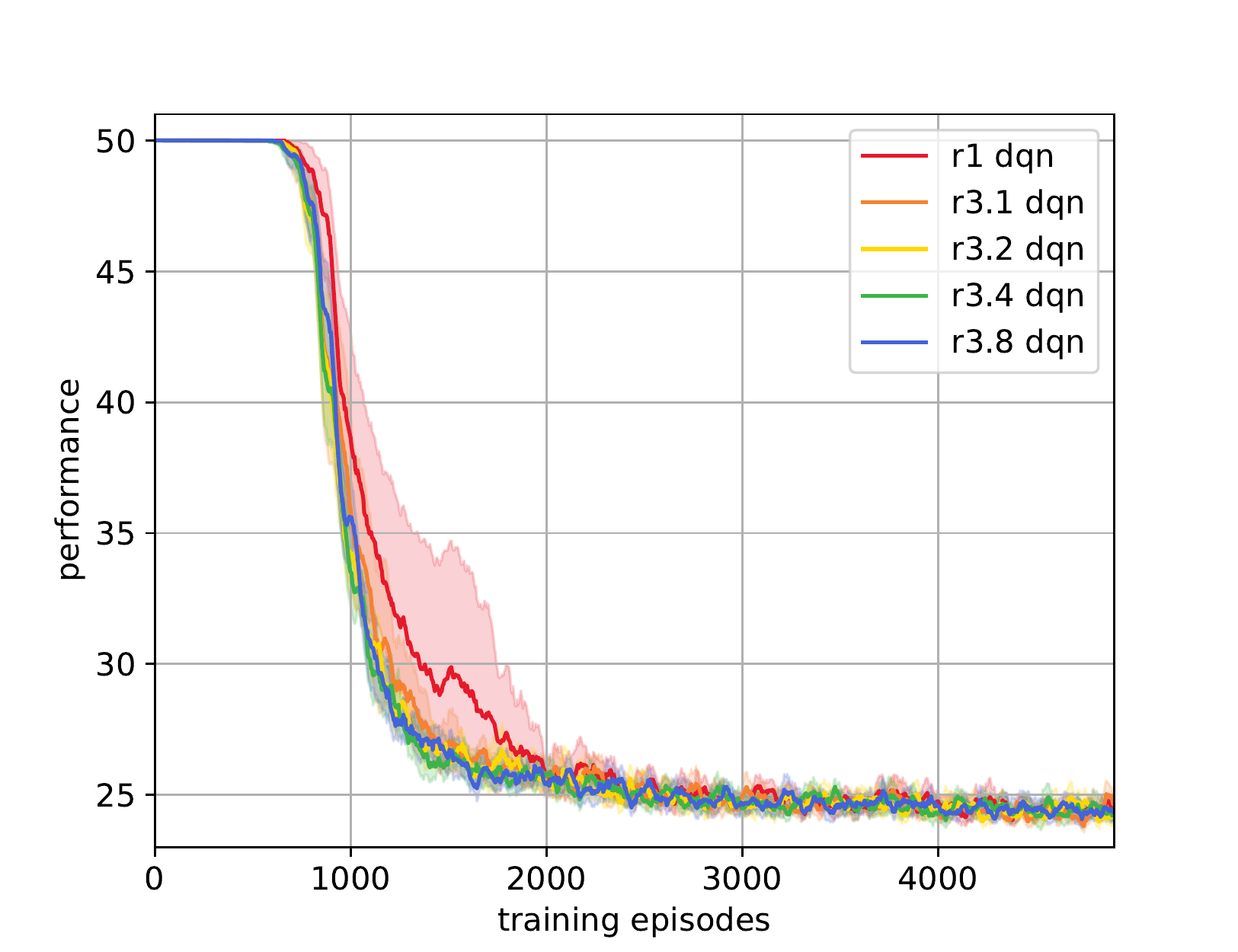}
    }
    \hfill
    \subfloat[performance when puni-\newline shing agent collisions\label{fig:ca_7}]{%
        \includegraphics[trim={0 0 30 30}, clip, width=0.24\textwidth]{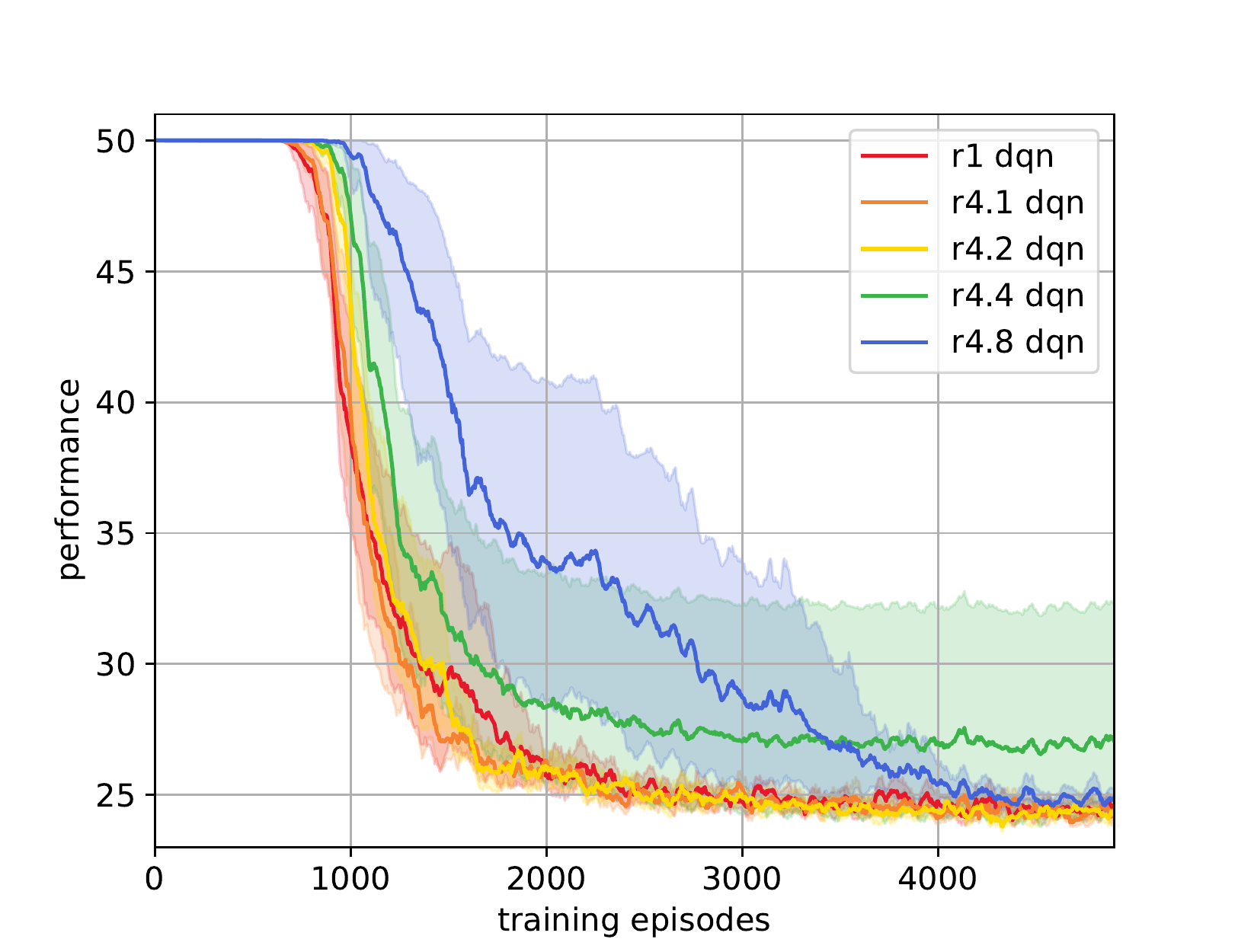}
    }
    \hfill
    \subfloat[performance overview of all static reward schemes\label{fig:ca_4}]{%
        \includegraphics[trim={0 0 30 30}, clip, width=0.24\textwidth]{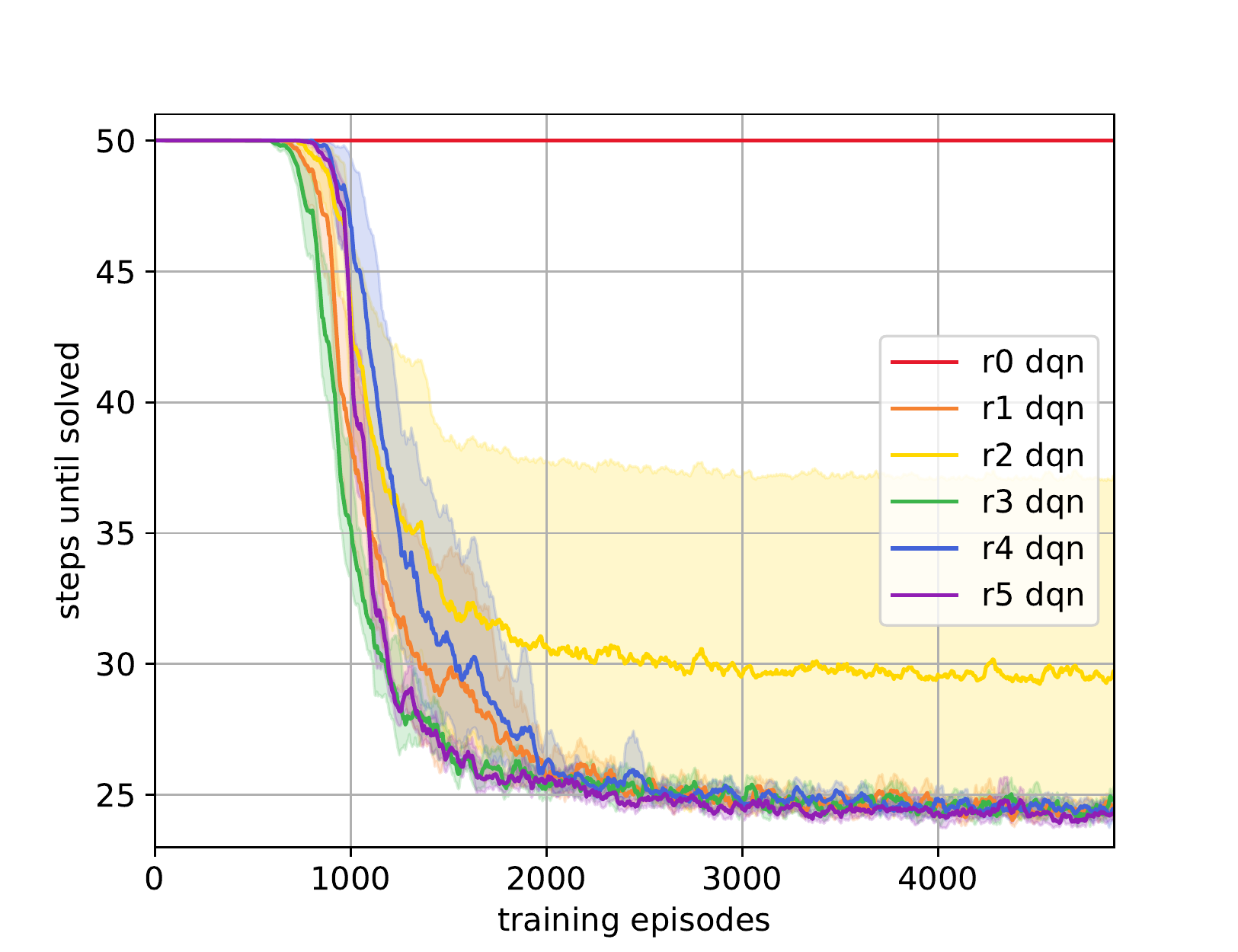}
    }
    \caption{Impact of isolated reward components (schemes \textit{r2}, \textit{r3} and \textit{r4}) on compliance and performance during training of 4 DQN agents. \textit{r0} is sparse with only functional components, \textit{r1} its dense counterpart. \textit{r5} contains all components of \textit{r1-r4}.}
    \label{fig:ca}	
\end{figure*}

\begin{figure}
	\captionsetup[subfigure]{justification=centering}
    \subfloat[compliance of 8 agents\label{fig:sa_0}]{%
        \includegraphics[trim={0 0 30 30}, clip, width=0.23\textwidth]{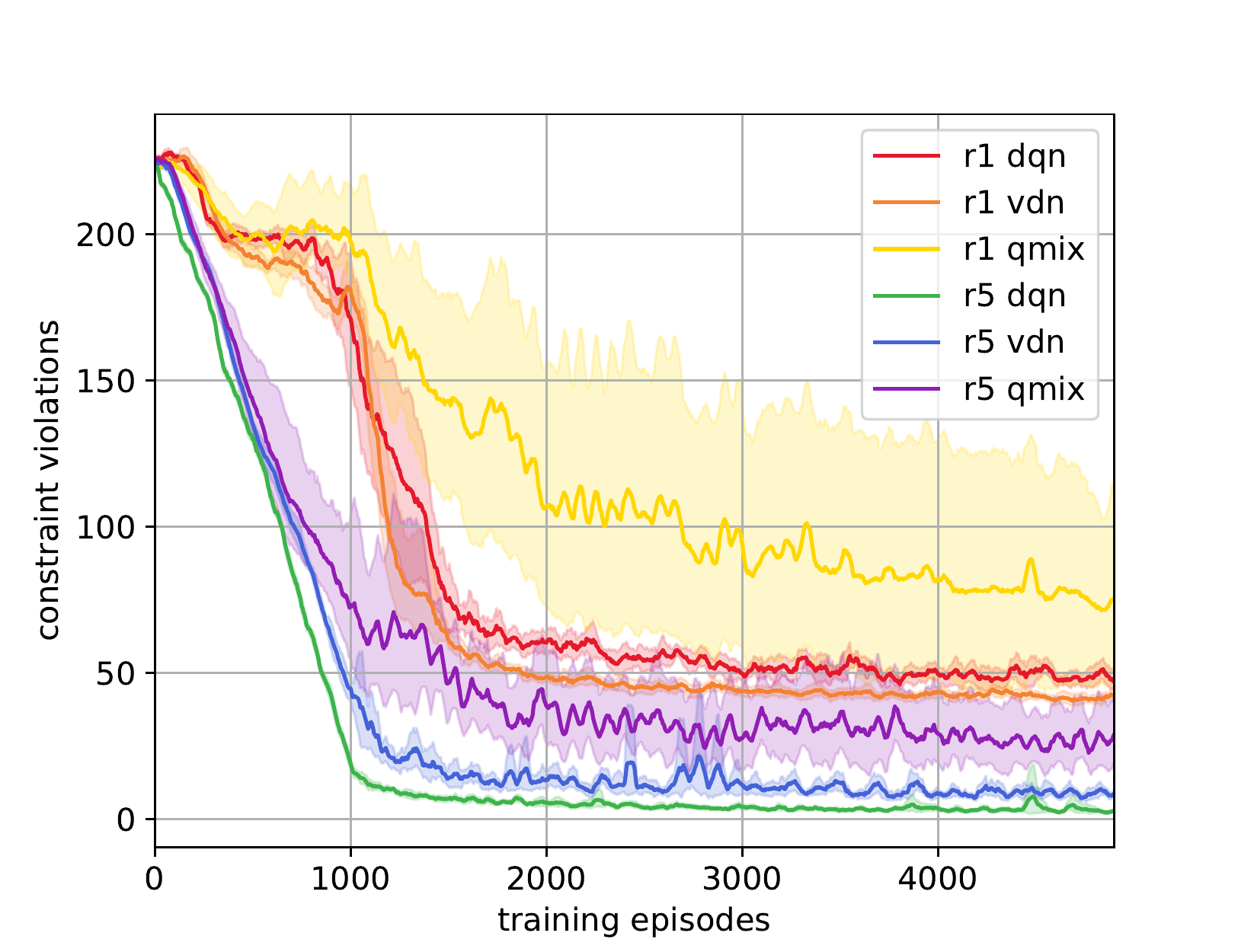}
    }
    \hfill
    \subfloat[performance of 8 agents\label{fig:sa_1}]{%
        \includegraphics[trim={0 0 30 30}, clip, width=0.23\textwidth]{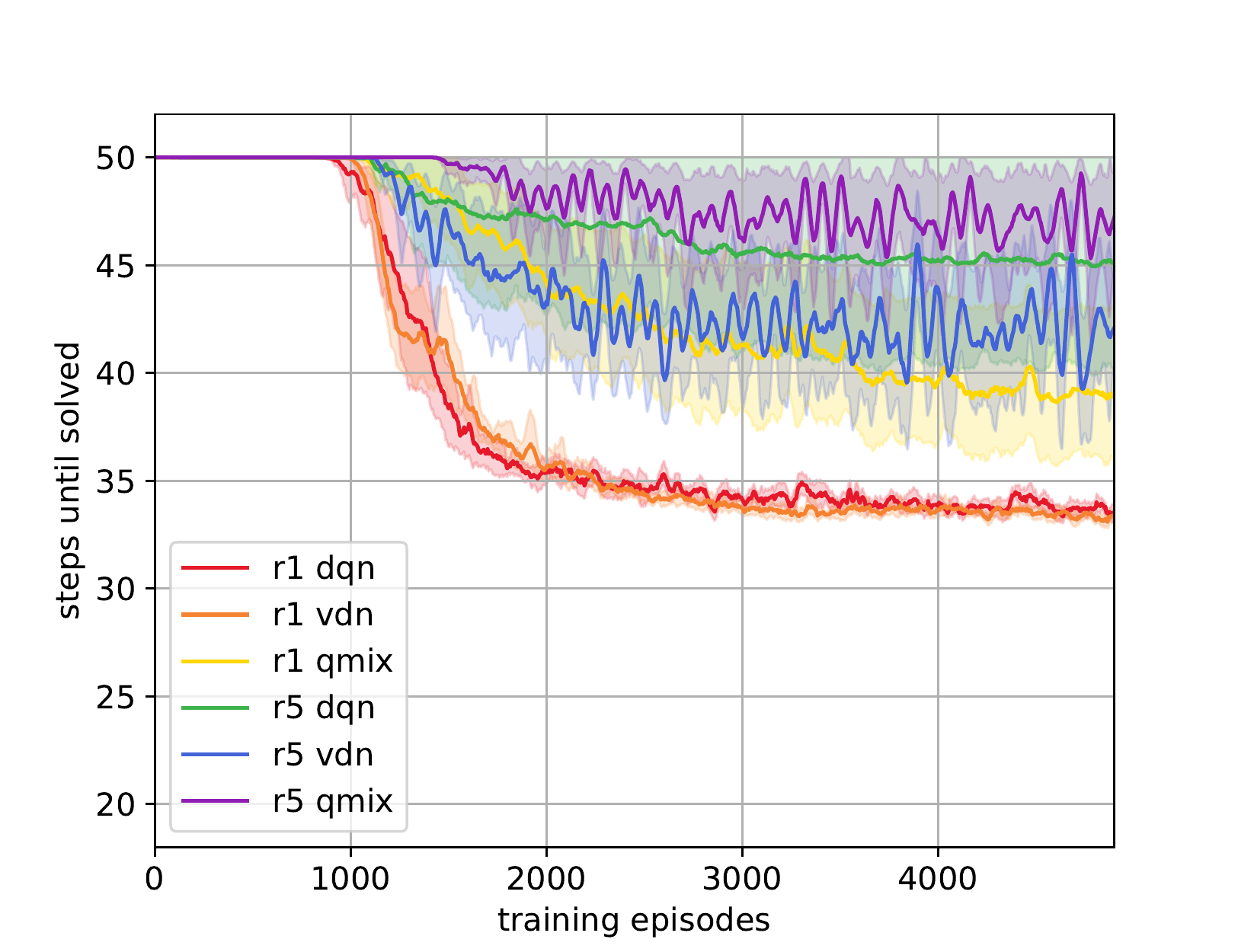}
    }
    \caption{Impact of combined reward components (scheme \textit{r5}) on compliance and performance during training of DQN, VDN and QMIX. \textit{r1} has only functional components.}
    \label{fig:sa}
\end{figure}

The results of the first scenario are depicted in Fig.~\ref{fig:ca}.
Excluding the sparse reward scheme \textit{r0} (see Fig.~\ref{fig:ca_0}, \ref{fig:ca_4}), convergence can be observed for all other reward schemes in all subfigures. Regarding compliance, wrong enqueueing and path violations can be minimized with small negative terms (see Fig.~\ref{fig:ca_1}, \ref{fig:ca_2}). In contrast, minimizing agent collisions requires greater negative potentials (see Fig.~\ref{fig:ca_3}). Regarding performance, punishing path violations makes no difference at all (see Fig.~\ref{fig:ca_6}) and punishing agent collisions slows down convergence only in case of big penalties (see Fig.~\ref{fig:ca_7}). However, punishing wrong enqueueing results in more steps until all tasks are finished. This effect becomes more significant with higher penalties (see Fig.~\ref{fig:ca_5}). Summing up the first scenario, reward scheme \textit{r5} notable decreases the overall constraint violations while the number of steps until all tasks are finished remains the same.

The results of the second scenario are shown in Fig.~\ref{fig:sa}. When scaled to $8$ agents, \textit{r5} also lowers the number of constraint violations with the impact differing per learning algorithm. Compared to DQN, VDN causes less violations with \emph{r1} and more violations with \emph{r5} (see Fig.~\ref{fig:sa_0}). Independent of the reward function, QMIX always causes more violations than DQN and VDN. Regarding performance, \textit{r5} causes an increased number of steps with compared to \textit{r1} with DQN, VDN and QMIX (see Fig.~\ref{fig:sa_1}). Overall in the second scenario, reward scheme \textit{r5} decreases the overall constraint violations but increases the number of steps until all tasks are finished.

\begin{figure*}
	\captionsetup[subfigure]{justification=centering}
    \subfloat[compliance: DQN, VDN\label{fig:gi_0_0}]{%
        \includegraphics[trim={0 0 30 30}, clip, width=0.24\textwidth]{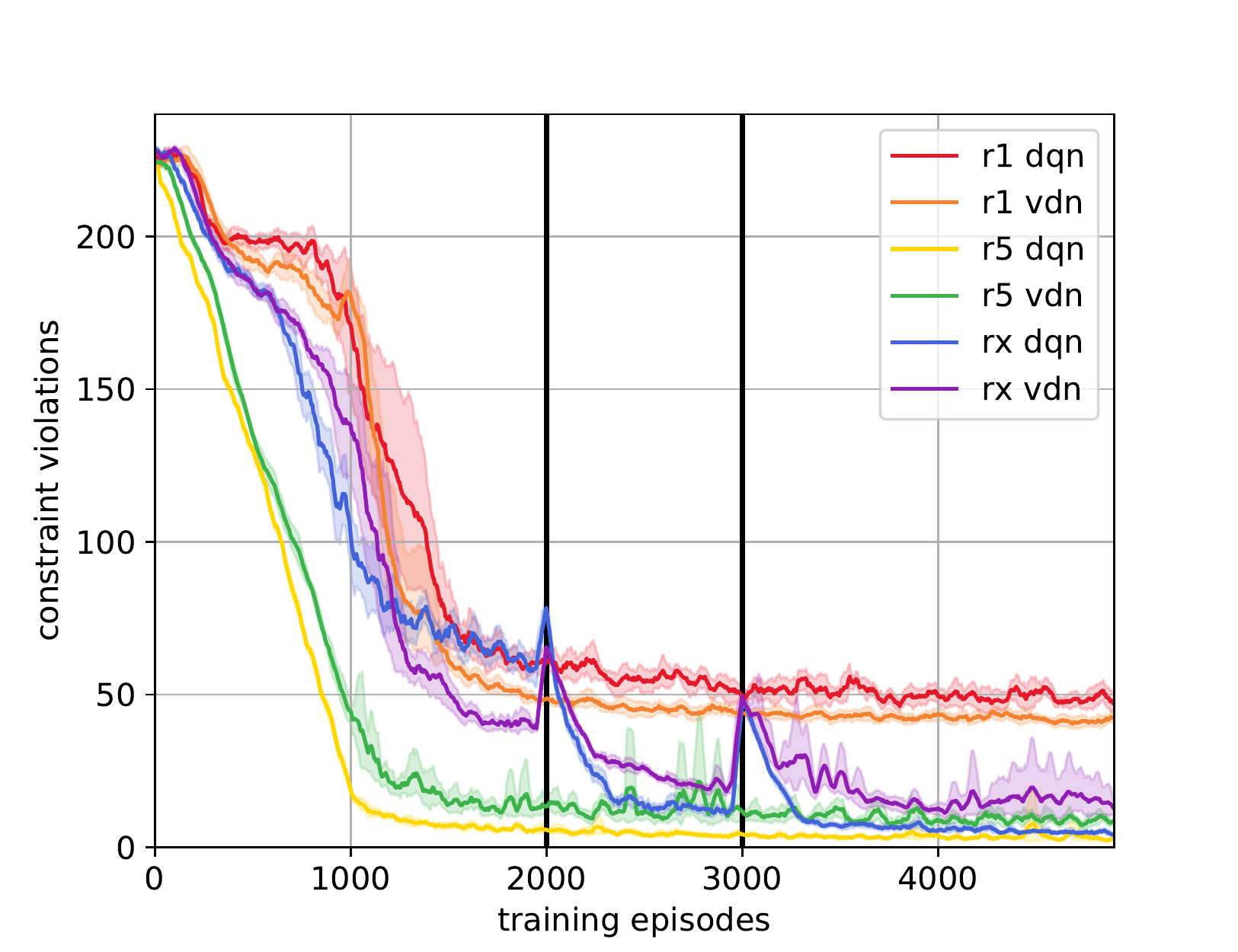}
    }
    \hfill
    \subfloat[compliance: DQN, QMIX\label{fig:gi_0_1}]{%
        \includegraphics[trim={0 0 30 30}, clip, width=0.24\textwidth]{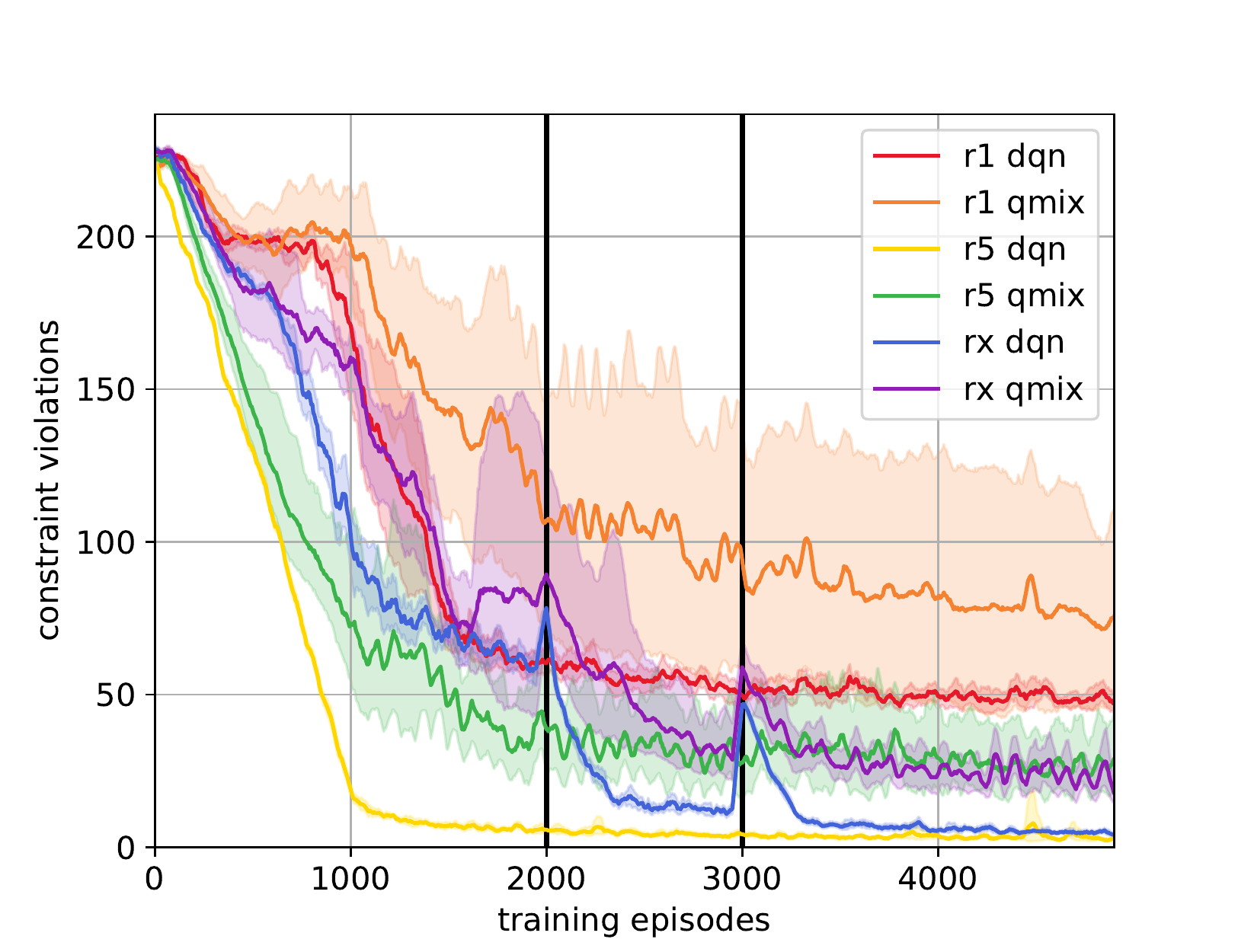}
    }
    \hfill
    \subfloat[performance: DQN, VDN\label{fig:gi_0_2}]{%
        \includegraphics[trim={0 0 30 30}, clip, width=0.24\textwidth]{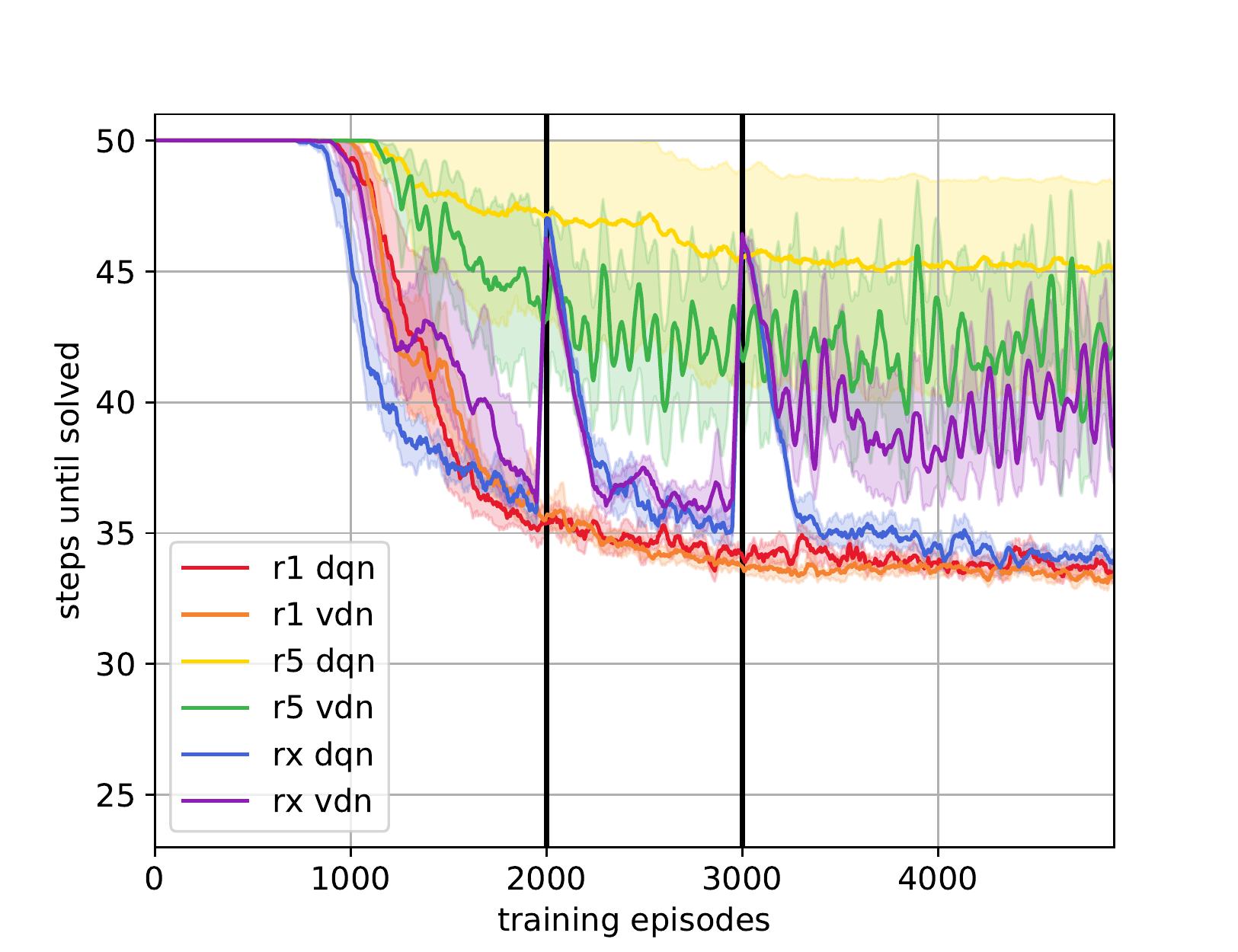}
    }
    \hfill
    \subfloat[performance: DQN, QMIX\label{fig:gi_0_3}]{%
        \includegraphics[trim={0 0 30 30}, clip, width=0.24\textwidth]{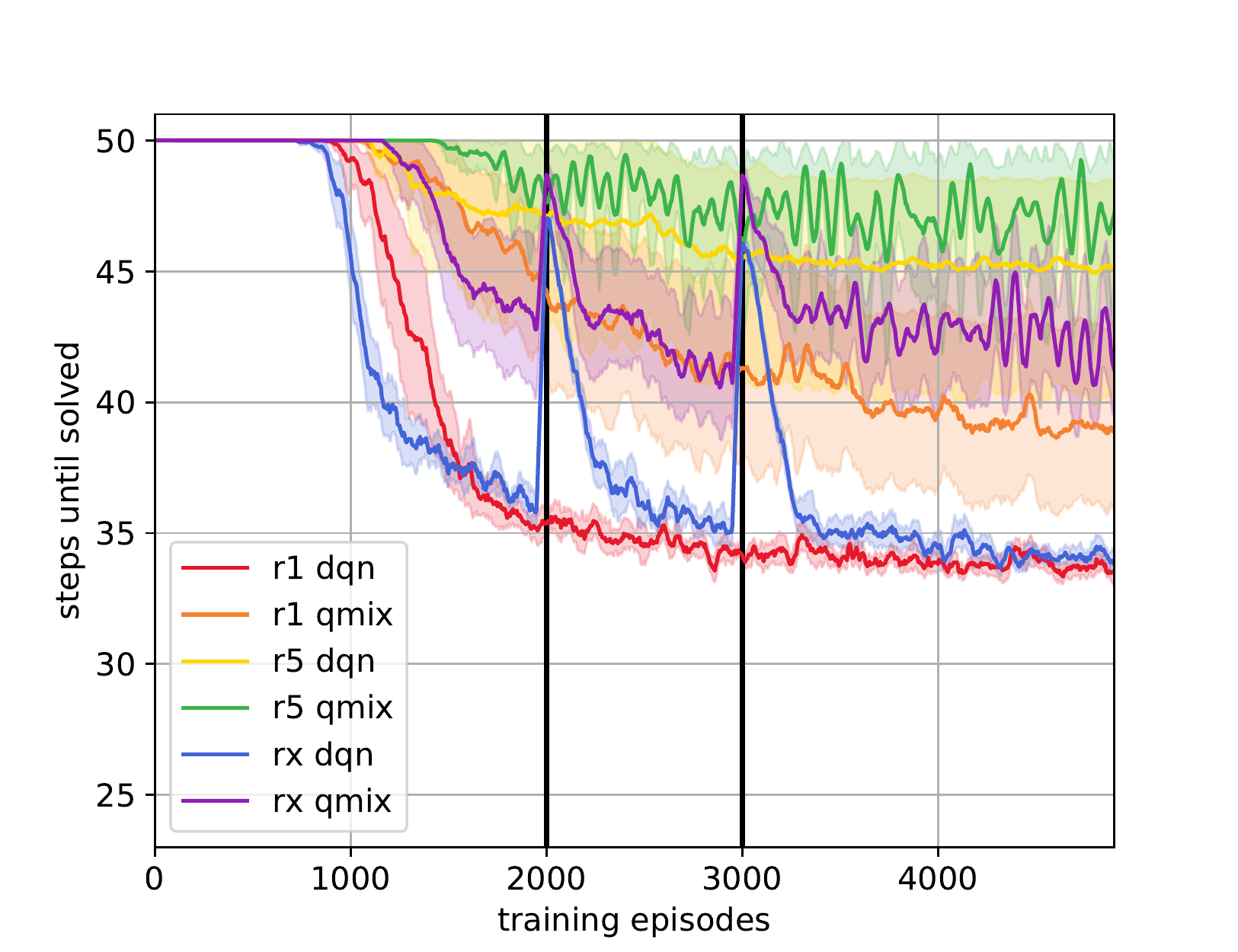}
    }
    \caption{Impact of gradually adding components at steps 2000 and 3000 via \textit{rx} on compliance and performance during training. \textit{r5} is a static scheme with all components, \textit{r1} is a static scheme with only functional components. 8 agents were trained with DQN, VDN and QMIX.}
    \label{fig:gi_0}
\end{figure*}

\begin{figure}
	\captionsetup[subfigure]{justification=centering}
    \subfloat[compliance of 6 agents\label{fig:gi_1_0}]{%
        \includegraphics[trim={0 0 30 30}, clip, width=0.23\textwidth]{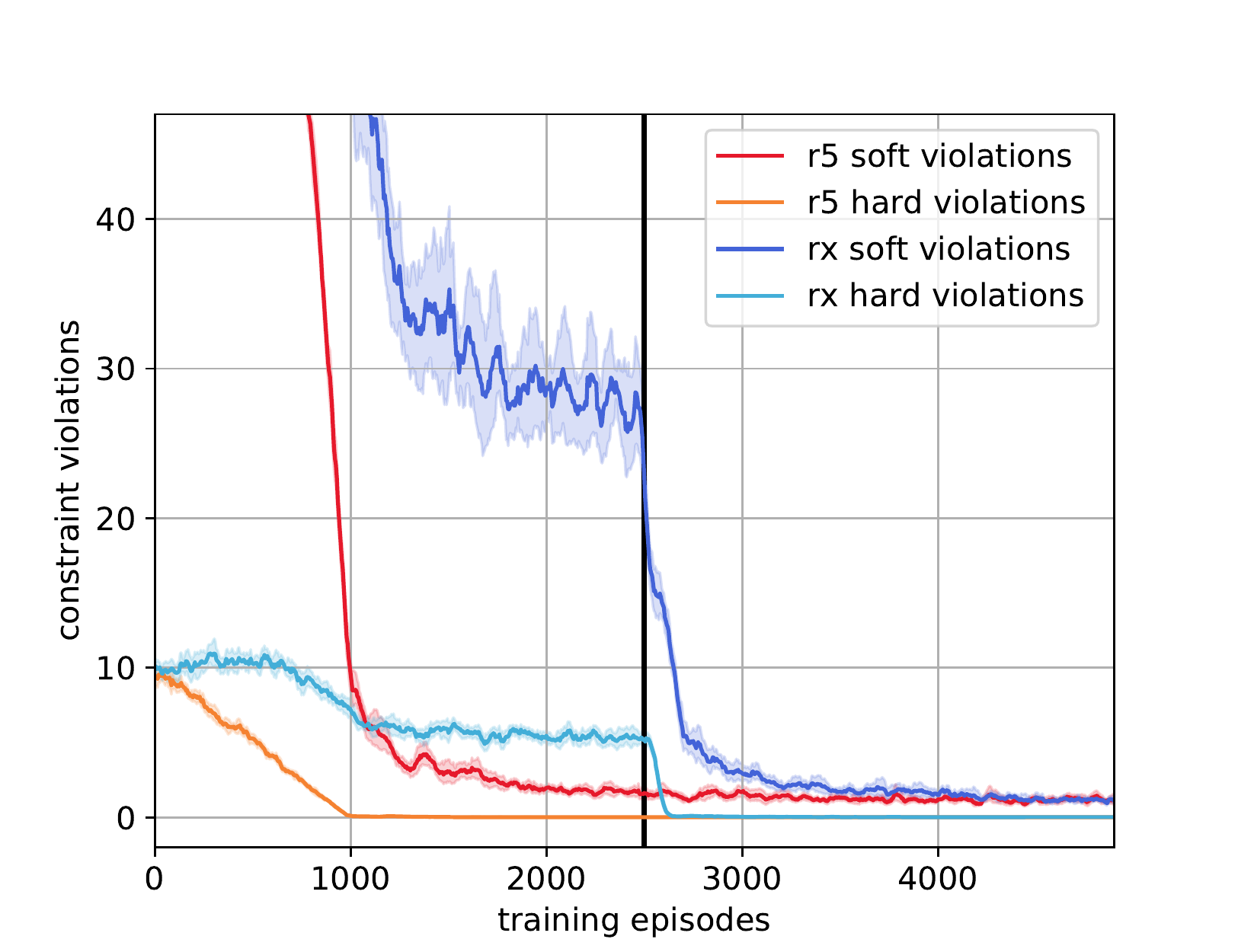}
    }
    \hfill
    \subfloat[performance of 6 agents\label{fig:gi_1_1}]{%
        \includegraphics[trim={0 0 30 30}, clip, width=0.23\textwidth]{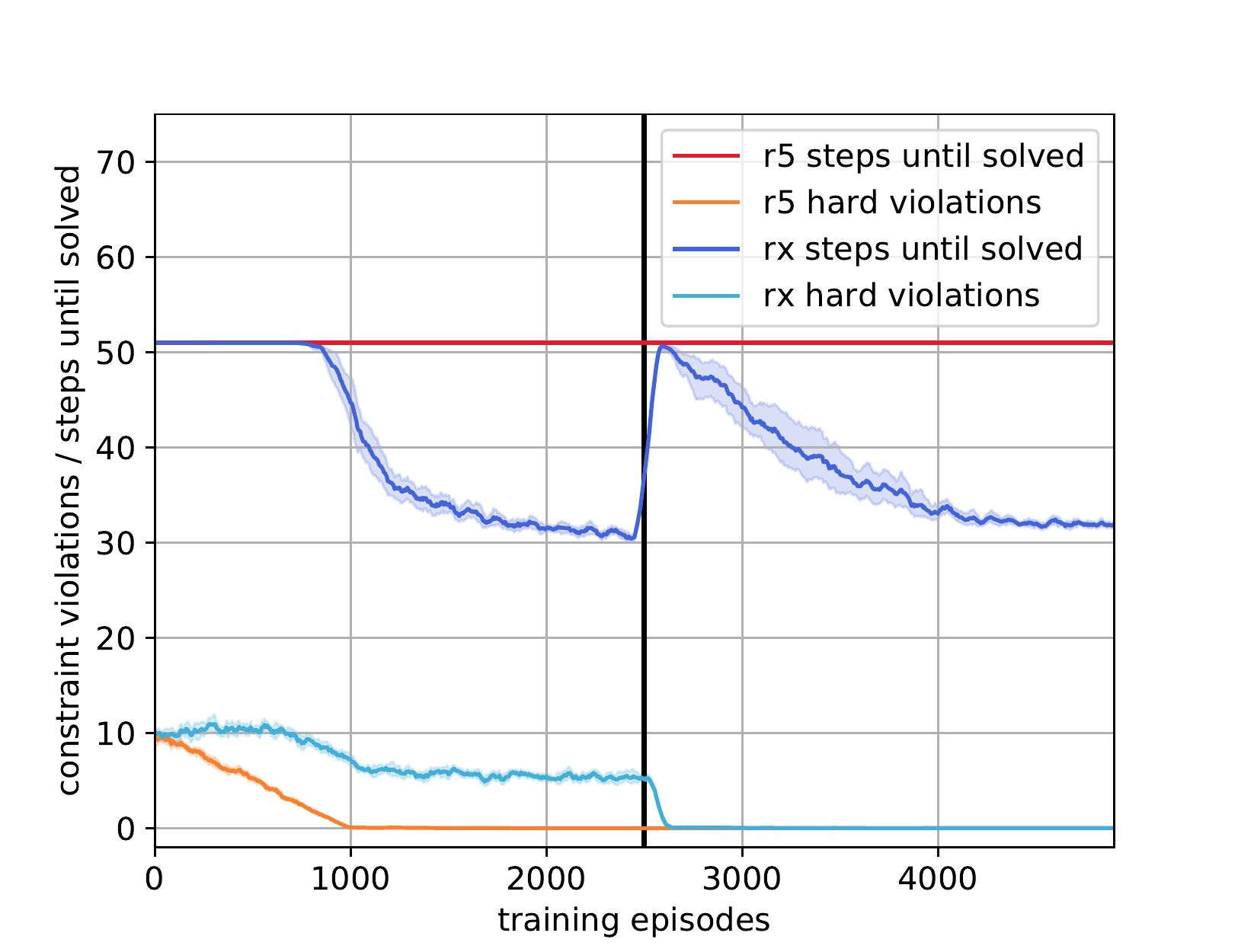}
    }
    \caption{Impact of gradually adding components at step 2500 via \textit{rx} during training of 6 DQN agents. Agents shall freeze in steps with active emergency signal (\emph{hard constraint}). All prior constraints are considered \emph{soft constraints}. \textit{r5} is a static scheme with the same components as \emph{rx} but weights them fully during the whole training.}
    \label{fig:gi_1}
\end{figure}

The results of the third scenario are depicted in Fig.~\ref{fig:gi_0}.
Note that in all subfigures, characteristic spikes at episode 2000 and 3000 are present in the data series of \textit{rx}. 
In terms of compliance, DQN and VDN show a reduced number of constraint violations with \emph{rx} compared to \textit{r1}, nearly as low as with \textit{r5} at the end of the training (see Fig.~\ref{fig:gi_0_0}). The same applies to QMIX, even though it converges slower (see Fig.~\ref{fig:gi_0_1}). Regarding performance, DQN with \textit{rx} approaches the level of \textit{r1} step-wish which is significantly lower than that of \textit{r5} (see Fig.~\ref{fig:gi_0_2}). Contrarily, \textit{rx} disrupts VDN's convergence, resulting in more steps than \textit{r1} and only sightly less than \textit{r5}. QMIX with \textit{rx} shows the same phenomenon (see Fig.~\ref{fig:gi_0_3}), although the final number of steps is between \textit{r1} and \textit{r5}.  Summing up the third scenario, reward scheme \textit{rx} leads to a step-wise decrease of constraint violations nearly to the level of \textit{r5}. However, agents with \textit{rx} solve the scenario faster than with \textit{r5}, sometimes as fast as with \textit{r1}.

The results of the fourth scenario are depicted in Fig.~\ref{fig:gi_1}. Due to a single reward adjustment, only one characteristic spike can be seen around episode 2500. While DQN with \emph{r5} succeeds to prevent hard  constraint violations and minimizes soft constraint violations with $6$ agents (see Fig.~\ref{fig:gi_1_1}), it fails to solve the environment in less than $50$ steps (see Fig.~\ref{fig:gi_1_1}). Contrary, \emph{rx} resolves the target conflict introduced by the emergency signals after the reward adjustment by maintaining adequate performance while minimizing all constraint violations.

\subsection{Discussion}

First of all, results indicate that the presented domain is suitable to benchmark practically relevant properties of a MAS. While this smart factory can be solved straightforwardly with up to $4$ agents, deploying $6$ and $8$ agents resulted in more constraint violations. Also, notably more steps were required and naive approaches struggled to finish their tasks. In such settings, we suppose agents to effectively compete for machines (to process items) and path segments (to navigate to the machines), leading to conflicts. Such scenarios require cooperative behavior between agents to perform well. Techniques restricting the action space cannot solve such scenarios alone as deadlocks would not necessarily be resolved.

Also, we observed sparse reward schemes such as \textit{r0} do not lead to convergence, which is not surprising. However, its decomposed counterpart \textit{r1} lead to solid performance throughout all scenarios. Although \emph{soft specification violations} in the smart factory are actions not solving the environment, \textit{r1} fails to completely minimize them in limited training time. Instead, reward schemes containing more specification components such as \textit{r5} turned out to increase compliance throughout all scenarios, which is considered a major insight.

Yet, depending on the learning algorithm and the number of agents, \textit{r5} partially increase the time until the environment was solved, especially when scaled to $8$ agents. Moreover, \textit{r5} failed to resolve the target conflict introduced with the hard constraints in Fig.~\ref{fig:gi_1}. Obviously, providing the whole specification to the agents from the beginning on does not necessarily lead to optimal behavior. Results of adaptive reward schemes such as \textit{rx} further show that starting with a basic reward scheme and gradually adding more components once the learner started to converge is capable of increasing specification compliance while maintaining performance. Conversely, some reward components may simply be unsuitable to begin the training with as they negatively affect exploration. As the learning algorithms also reacted differently to \textit{rx}, timing when to add components and the correct value is assumed to be substantial. Such side-effects of reward shaping have also been reported in the literature \cite{dyk14}.




\section{Conclusion and Future Work}
\label{sec:conclusion}

In this work, we considered the problem of specification compliance in MARL.

We introduced an involved multi-agent domain based on a smart factory setting. We translated the system's goal specification into a shaped reward function and analyzed how the system's non-functional requirements can be modeled by adding more terms to that reward function. Besides the raw performance, we also evaluated specification compliance in RL on a multi-agent setting.

While simple shaped rewards, which weight only functional requirements like the task rewards or the step cost, can lead to agents that are able to achieve the basic goal, our results show that they still have a high tendency to violate the non-functional requirements, which could be harmful for industrial or safety-critical domains.

Our approach to explicitly translating these requirements into a shaped reward function was shown to still enable agents to solve the global goal, while being able to consider these additional requirements, making them specification-aware.

In accordance with other results~\cite{cl09,gupta2017cooperative}, we found an inherent benefit due to gradually applied rewards where reward function and scenario become increasingly more complicated.

An immediate generalization of our experiments would be to replace the hand crafted shaping and scheduling    with some kind of auto-curriculum mechanism. This would allow the reward functions to adjust themselves in direct response to the agents' learning progress \cite{jcw19,llm19} but focusing on non-functional objectives. Such techniques have also been employed for adversarial learning~\cite{lowd2005adversarial}, but our results sketch a path on how to implement reward engineering as an adversary given a fixed specification.

As we now only considered a cooperative setting, it seems natural to expand our study to groups of self-interested agents with (partially) opposing goals. As these all have their own reward signal, deriving reward functions from a shared specification for the whole systems becomes dramatically more difficult (at least doing so manually). However, especially in industrial applications, ensuring safety between parties with different interests is all the more crucial and formulating the right secondary reward terms to ensure ``fair play'' might allow for even greater improvement in the whole system's performance than it does for the cooperative setting.

Eventually, we would suspect that results from our evaluation can be applied back to the formulation of the original specification. Requirements that are not needed within the reward function might not rightfully belong in the specification. This way, the usually human-made specification can be improved via the translation to a reward function and the execution of test runs of RL.

\bibliographystyle{apalike}
{\small
\bibliography{paper}}

\end{document}